# AI Planning: A Primer and Survey (Preliminary Report)

**Dillon Z. Chen**[1], **Pulkit Verma**[2], **Siddharth Srivastava**[3], **Michael Katz**[4], **Sylvie Thiébaux**[1,5]

[1]LAAS-CNRS, University of Toulouse
[2]Massachusetts Institute of Technology
[3]Arizona State University
[4]IBM Research
[5]The Australian National University

**Abstract**

Automated decision-making is a fundamental topic that spans multiple sub-disciplines in AI: reinforcement learning (RL), AI planning (AP), foundation models, and operations research, among others. Despite recent efforts to "bridge the gaps" between these communities, there remain many insights that have not yet transcended the boundaries. Our goal in this paper is to provide a brief and non-exhaustive primer on ideas well-known in AP, but less so in other sub-disciplines. We do so by introducing the classical AP problem and representation, and extensions that handle uncertainty and time through the Markov Decision Process formalism. Next, we survey state-of-the-art techniques and ideas for solving AP problems, focusing on their ability to exploit problem structure. Lastly, we cover subfields within AP for learning structure from unstructured inputs and learning to generalise to unseen scenarios and situations.

## 1 Introduction

Reinforcement Learning (RL) and AI Planning (AP) are two subfields of AI that cover autonomous decision-making: the act of mapping situations to actions in order to satisfy a specified objective such as maximising a reward signal and/or achieving a goal condition. RL involves interacting with *unknown environments* guided by rewards to discover the best actions to take (Sutton and Barto 1998), while conversely, AP reasons over *structured models* represented in formal and compact languages to solve long horizon problems with sparse rewards (Geffner and Bonet 2013).

A core insight from AP lies in leveraging structure in world models for allowing agents to make decisions more efficiently and effectively than if we were to treat the world as a black box. AP problems are represented by compact and symbolic models which yield a plethora of theoretical tools and powerful, domain-independent algorithms. Traditional AP approaches involve manually designing world models with declarative languages that capture structures inherent to a domain, with examples including the early STRIPS formalism (Fikes and Nilsson 1971) to the more recent PDDL (McDermott et al. 1998; Haslum et al. 2019) and RDDL (Sanner 2010) formalisms. In this paradigm, it is the engineer who recognises the structure in the world, and then the planner that effectively leverages that structure.

Another line of work exploits learning to automatically abstract and discover structure in problems from raw inputs for making decisions computationally feasible (Konidaris 2019). On top of using learning to make planning feasible the discovery of structured models, we can also learn to make planning fast. By making use of structural priors in AP formalisms, learning approaches also exhibit powerful generalisation capabilities by scaling to or encoding compact solutions of unseen problems of arbitrary size.

To organise the various AP topics we aim to cover, we structure this survey paper as follows.

- In Sec. 2, we formalise the common RL problem setup as an MDP problem from which we introduce common assumptions and formalisms of the AP problem.

This involves classifying different levels of access an agent has to an MDP. On the RL-side of the spectrum, agents have little to no information of the MDP and have no ability to undo their moves. On the AP-side of the spectrum, agents have complete access to the MDP which is represented with compact and structured models.

- In Sec. 3, we provide a primer into the AP problem and survey state-of-the-art methodologies in the field which exploit the AP assumptions.

We formalise structured AP problems using first-order logic and the closed world assumption for compactly encoding combinatorially large world models. We also cover expressive extensions for handling uncertainty, time, and environmental processes. Then we highlight key ideas and themes for exploiting such structure, while also drawing some analogies to common RL topics. We conclude with pointers to resources for getting started in AP.

- In Sec. 4, we cover subfields of AP involving learning. This involves both learning structure for performing AP, as well as learning from structure for improving AP.

More specifically, we cover AP techniques which can exhibit generalisation capabilities. This involves the ability to handle more than one problem at a time, which we now classify in terms of an algorithm's level of access to *sets of* MDPs. The fields we cover include Generalised Planning (GP), which entails compactly representing solutions to infinite sets of problems, Learning for Planning (L4P), which involves learning to plan quickly from training data and structure, and Learning Planning Models (LPM), which involves mapping arbitrary, unstructured MDPs to AP models for more efficient planning.

## 2 MDPs: Access and Structure

We begin with a very general problem formulation. Consider a Markov Decision Process (MDP) $\mathcal{M} = \langle \mathcal{S}, \mathcal{A}, \mathcal{T}, \mathcal{R}, \mathcal{G} \rangle$ where $\mathcal{S}$ is a set of states, $\mathcal{A}$ is a set of actions, $\mathcal{T} : \mathcal{S} \times \mathcal{A} \times \mathcal{S} \to [0, 1]$ is a transition function, where $\mathcal{T}(s, a, \cdot)$ is a probability distribution, $\mathcal{R} : \mathcal{S} \times \mathcal{A} \times \mathcal{S} \to \mathbb{R}$ is a reward function, and $\mathcal{G} \subseteq \mathcal{S}$ is a set of terminal states.[1] MDPs can be extended to include partial observability; we assume full observability for the sake of a focused discussion. A solution to an MDP is a deterministic policy $\pi : \mathcal{S} \to \mathcal{A}$. The value function $V^\pi : \mathcal{S} \to \mathbb{R}$ for a policy $\pi$ gives the expected cumulative reward for executing the policy starting from a state. This function can be defined with a Bellman equation:

$$V^\pi(s) = \begin{cases} 0 \text{ if } s \in \mathcal{G}, \text{ and otherwise,} \\ \sum_{s' \in \mathcal{S}} \mathcal{T}(s, \pi(s), s') \left( \mathcal{R}(s, \pi(s), s') + V^\pi(s') \right). \end{cases}$$

A policy $\pi^*$ is optimal if $V^{\pi^*}(s) = \max_\pi V^\pi(s)$ for all $s \in \mathcal{S}$. A policy is proper if it is guaranteed (with probability 1) to reach a terminal state. It is sometimes difficult to guarantee optimality or properness, in which case we prefer policies that obtain high expected cumulative rewards with respect to an initial state distribution.

What level of *access* does an agent have to its MDP? This is an important question on which prior works differ. Sutton and Barto (1998) provides a binary classification by describing MDPs as either *distribution models*, also known as model-based RL, where $\mathcal{R}$ and $\mathcal{T}$ are known, or *sample models*, also known as model-free RL, where $\mathcal{R}$ and $\mathcal{T}$ must be sampled from. We further refine the hierarchy of access:

1. **Continuing:** The agent has no access to $\mathcal{T}$ or $\mathcal{R}$ and it "only lives once." The agent exists at some state $s$, executes an action $a$, and observes the next state $s'$ and reward $r$, repeating this process until $s \in \mathcal{G}$ (e.g., death).

2. **Episodic:** The agent still has no access to $\mathcal{T}$ or $\mathcal{R}$, but when a sink state is reached, the world resets to a state sampled from an initial state distribution. This level of access is very common in reinforcement learning (e.g., in OpenAI Gym (Brockman 2016)).

3. **Generative:** The agent has knowledge of $\mathcal{R}$, but only a *generative* or *sampling* access to $\mathcal{T}$, such as with a simulator. For any $s \in \mathcal{S}$ and $a \in \mathcal{A}$, the agent can sample $s' \sim \mathcal{T}(s, a)$. This access level is common in cases where the set of next possible states is too large to enumerate.

4. **Analytic:** The agent has direct access to $\mathcal{T}$ and $\mathcal{R}$ as in a distribution model MDP. This means that for every $s \in \mathcal{S}$ and $a \in \mathcal{A}$, the agent knows exactly the outputs of the functions $\mathcal{T}(s, a, \cdot)$ and $\mathcal{R}(s, a, \cdot)$.

5. **Structured:** The agent has direct access to $\mathcal{T}$ and $\mathcal{R}$ as in an analytic MDP, and furthermore the agent has access to *structured representations* of these models that may be leveraged for efficient and effective decision-making.

---

[1] This is an *indefinite-horizon* MDP. Infinite-horizon MDPs with discount factors can be converted into equivalent indefinite-horizon MDPs. Indefinite-horizon MDPs should also have *proper policies* to be well-defined; see Mausam and Kolobov (2012, Sec. 2.4).

We focus in this paper on what is meant by structured representations, and methods that make use of such structure in many different forms. Later in the paper, we will also discuss how lower levels of MDP access can effectively be *upgraded* to structured access via learning.

## 3 Planning

In this section, we provide a brief overview of formalisms and algorithms for AP as structured representations of MDPs. We conclude with pointers to more comprehensive references and repositories for interested readers to get started with planning.

### 3.1 AP Representations

AP practitioners formalise decision-making problems as structured MDPs, which compactly represent large worlds, with the aid of first-order logic and the closed world assumption. We begin by providing a minimal formalism of structured MDPs by covering how one is able to represent states and transitions in a factored form. The formalism presented in this section is commonly referred to as the *classical planning* setup in which problems are discrete, fully observable and deterministic. However, the presented ideas can generalise to involve probabilities, as well as further extensions of the presented MDP formalism, which we cover in Sec. 3.3.

**States** The necessary recipe components for constructing states in a planning task include (1) a set of predicates $\mathbf{P}$, and (2) objects $\mathbf{O}$. A predicate takes the form $p(x_1, \ldots, x_n) \in \mathbf{P}$ where $p$ is simply a predicate symbol, and the $x_i$'s are its arguments. The arity of a predicate is the number of arguments it takes. The set of objects $\mathbf{O}$ is a set of constant symbols. With these two ingredients, one can define *propositions*, which are obtained from predicates by assigning objects from $\mathbf{O}$ to all their arguments. The process of assigning objects to predicate arguments is called grounding. For example, `in(x,y)` is a predicate with arguments $x$ and $y$, and can be grounded into a proposition `in(dog,bedroom)`, which means that the dog is in the bedroom.

A state in a planning task is simply a set of propositions. AP representations follow the closed-world assumption, which states that the propositions in the state represent all true facts, and anything else not in the state is assumed false. For example, suppose that our world also contains the `dirty` predicate and the current state is

{`in(dog,bedroom)`, `hungry(dog)`, `brown(dog)`}.

Given that `dirty(dog)` is not present in the state, we know that the dog is not dirty.

**Goals** AP problems are goal-oriented, as opposed to reward-oriented in the common RL problem treatment. This treatment of decision-making problems can be represented in our MDP formalism by treating terminal states as goal states. A goal condition $\mathbf{G}$ in a planning task is defined as a propositional formula, and a state $s$ is a goal state if its closed-world interpretation satisfies the formula $\mathbf{G}$. Thus, we can implicitly represent the set of all goal states $\mathcal{G}$ in an MDP without enumerating all of them. For example, we may have a goal condition ¬`hungry(dog)` which can describe all goal states regardless of where the dog is located.

**Transitions** Transitions in AP are defined from a set of action schemata $\mathbf{A}$, which similarly to predicates, can be grounded from task objects to induce a high-order polynomial number of actions and transitions. In the goal-oriented setting, such actions exhibit negative rewards in the reward maximisation setting in order to represent action costs.

An action schema in a planning task is a tuple $\mathbf{a} = \langle \arg(\mathbf{a}), \text{pre}(\mathbf{a}), \text{add}(\mathbf{a}), \text{del}(\mathbf{a}) \rangle$, where $\arg(\mathbf{a})$ denotes a set of arguments and $\text{pre}(\mathbf{a})/\text{add}(\mathbf{a})/\text{del}(\mathbf{a})$ are sets of predicates from $\mathbf{P}$ instantiated with arguments from $\arg(\mathbf{a})$ representing the preconditions (pre) that must be true for the action to be executable, and its positive (add) and negative (del) effects. For example, we can define a `move` schema by

$$\arg(\mathbf{a}) = \{x, l_1, l_2\}, \quad \text{pre}(\mathbf{a}) = \{\text{in}(x, l_1)\},$$
$$\text{add}(\mathbf{a}) = \{\text{in}(x, l_2)\}, \quad \text{del}(\mathbf{a}) = \{\text{in}(x, l_1)\}.$$

An action schema $\mathbf{a}$ can be grounded into an action $a$ by assigning objects from $\mathbf{O}$ to arguments in $\arg(\mathbf{a})$. Each action corresponds to a set of transitions. More specifically, for each state $s$ and action $a$, we say that $a$ is applicable in $s$ if $\text{pre}(a) \subseteq s$, in which case we define the successor state by $s' = (s \setminus \text{del}(a)) \cup \text{add}(a)$. Preconditions and action effects implicitly induce the probability values of transitions. More specifically, we have $\mathcal{T}(s, a, s') = 1$, and $\mathcal{T}(s, a, s'') = 0$ for all $s'' \neq s'$. Furthermore, if $a$ is not applicable in $s$, we have $\mathcal{T}(s, a, t) = 0$ for all states $t$.

For example, the defined action schema `move` induces the action `move(dog, bedroom, kitchen)` when we assign the schema arguments to the objects `dog`, `bedroom`, and `kitchen`. This action can be applied in the previously mentioned state to create with probability 1 a new state

{in(dog, kitchen), hungry(dog), brown(dog)}.

### 3.2 Benefits of Structure

The presented AP representation for modelling MDPs provides several benefits for efficient decision-making.

**Attention** The use of predicate logic in AP allows us to view states as databases from which one may make queries to determine transitions and relevant information for the goal. Suppose we extend our example state to encode other true but irrelevant information about the world:

{in(dog, bedroom), hungry(dog), brown(dog),
 dirty(bedroom), likes(dog, bedroom),
 can_make(dog, sandwich), in(bed, bedroom)}.

The previous move action is still applicable in this state and due to the implicit frame axiom of the AP formalism, applying the action does not change the additional, irrelevant facts. One can draw an analogy of the database view of AP to the ML concept of attention where one focuses only on the necessary information in the input for making decisions. Indeed, AP approaches leverage this relevance insight (Muise, McIlraith, and Beck 2012; Corrêa et al. 2020; Silver et al. 2021) for achieving effective planning performance in complex environments.

**Automation** AP can be viewed as a declarative programming paradigm where problems are specified as a program in a planning language such as PDDL. The primary advantage from this viewpoint is that one does not have to write a solver for a problem and only a model of the problem. This can save time and effort required for developing solvers and also making changes to the problem. Furthermore, by assuming standardised input languages for planning, AP researchers have developed highly optimised planners.

**Complexity** The usage of predicate logic also allows us to compactly represent combinatorially large state spaces. For example, if we have a predicate of arity $k$ and there are $n$ objects, then the maximum number of possible propositions we can create is $n^k$. However, thanks to the closed-world assumption, we do not have to explicitly mention all false propositions so states are generally very small and compactly represented. The case is similar for action schemata. Indeed the compactness of the planning task formalised as above is reflected in the result that solving a planning task represented in this way is EXPSPACE-complete (Erol, Nau, and Subrahmanian 1995).

### 3.3 Planning Extensions

The previous section identified the key ingredients commonly used in AP research from and for which a plethora of powerful algorithms have been developed. Although we only covered the classical planning setup so far, there exist various AP extensions which bootstrap off similar ideas for handling more complex features. Without going into formal details, we present pointers to such AP extensions.

**Uncertainty** *Probabilistic Planning* (Mausam and Kolobov 2012) extends the classical planning setup by allowing for probabilistic transitions, thus encapsulating the presented MDP formalism. This has been done in PPDDL (Younes and Littman 2004) by extending action schemata with sets of effects with associated probabilities. RDDL (Sanner 2010) is another language which can encode factored MDPs as well as specifying common probability distributions of actions. It differs from PDDL by providing a fluent-centric representation of actions and can also be viewed as a Dynamic Bayesian Network.

*Nondeterministic Planning* (Cimatti et al. 2003) is a variant of probabilistic planning where probabilities for transitions are not given or unknown. *Conformant Planning* (Smith and Weld 1998) is the problem of computing plans in partially observable environments that are guaranteed to succeed with probability 1 or greater than a specified threshold (Domshlak and Hoffmann 2006, 2007).

**Time and Numerics** PDDL 2.1 (Fox and Long 2003) extends classical planning by introducing both time and numerics. *Numeric Planning* introduces functions and numeric expressions, allowing the representation of numeric state variables capturing e.g. resources, physical properties, and plan metrics. *Temporal Planning* extends the definition of actions and plans to allow for durative and concurrent actions, as well as deadlines and temporal synchronisation.

**Beyond Agentic Setups** The classical planning setup is agent-centric, where environmental processes are modelled as if they were performed by the agent. *Hybrid Planning* (Fox and Long 2006), formalised in the PDDL+ language, introduces continuous variables as well as the abil-ity to model continuous processes and events for representing environmental changes. *Multi-Agent Planning* (Torreño et al. 2018) has been represented in MA-STRIPS (Brafman and Domshlak 2008) for which proper encodings lead to more efficient planning as opposed to modelling multiple agents with classical planning.

### 3.4 State-of-the-Art from Structure

Despite the computational complexity of solving planning tasks, AP researchers have developed a number of powerful techniques and solvers for tackling them. In this section, we focus on the main methodologies underlying the former, and also draw ties to similar themes from the RL literature. The presented techniques originate from classical planning but the underlying ideas generalise to various extensions covered in Sec. 3.3 such as probabilistic planning.

We begin by describing components of heuristic search, a state-of-the-art technique for planning. The components involve the derivation of (1) domain-independent heuristic functions for use with (2) heuristic search algorithms. This approach draws analogies to state-of-the-art RL methodologies involving value function approximation and search (Silver et al. 2016; Sutton 2019). Next, we cover a wide variety of other planning techniques falling under the umbrella of term of (3) problem decomposition. This subsection draws analogies to the diverse field of Hierarchical RL (HRL). More specifically, both HRL and decomposition techniques in planning involve breaking down larger difficult problems into smaller, more manageable subproblems.

**Heuristic Functions**   Some RL algorithms employ approximate value functions to estimate the maximum expected reward achievable from a given state over infinite horizons with discounts or finite horizons to help compute optimal policies. Contrarily, some AP algorithms use heuristic functions which estimate the optimal cost-to-go to a goal state for computing solutions. Formally, a heuristic is a function on states $h : \mathcal{S} \to \mathbb{R} \cup \{\infty\}$ where $\infty$ is used to represent deadend states, states that cannot reach the goal state. The optimal heuristic $h^*$ returns the optimal plan cost from a state to a goal, and a heuristic $h$ is admissible if $h(s) \leq h^*(s), \forall s \in \mathcal{S}$. Admissible heuristics used in the A$^*$ search algorithm are guaranteed to return optimal solutions, while inadmissible heuristics are used with search algorithms such as Greedy Best First Search (GBFS) for returning solutions quickly, but not necessarily optimal.

Heuristic functions in planning are often synthesised by approximating $h^*$ through solving easier relaxations of the original task. The *delete-relaxation* of a planning task involves removing delete effects of all actions and leads to the derivation of widely-known planning heuristics including the inadmissible FF (Hoffmann and Nebel 2001) and the admissible LM-cut heuristic (Helmert and Domshlak 2009).

There also exists a suite of heuristics arising from solving *abstractions* of the original task. Early works include extending Pattern Database (PDB) heuristics (Culberson and Schaeffer 1996, 1998) to planning (Edelkamp 2002; Haslum et al. 2007), which in turn have later been generalised to Merge and Shrink (M&S) abstractions (Helmert, Haslum, and Hoffmann 2007). Cartesian abstractions (Seipp and Helmert 2013, 2018) also generalise PDB heuristics and provide more efficient abstraction refinements than M&S.

*Cost partitioning* (Katz and Domshlak 2008) provides an approach for adding multiple heuristics into one in an admissible way by partitioning the cost of each action among these heuristics. It has been used as a theoretical tool to compare different classes of heuristics (Helmert and Domshlak 2009) and has also been extended in various fashions for constructing strong admissible heuristics (Pommerening, Röger, and Helmert 2013; Pommerening et al. 2014; Seipp and Helmert 2014; Seipp, Keller, and Helmert 2020).

*Novelty* heuristics and width-based algorithms (Lipovetzky and Geffner 2012, 2017; Katz et al. 2017) make use of the structured representation of planning tasks for balancing exploration and exploitation for suboptimal planning through preferring states that exhibit subsets of facts that have not been seen before.

**Search Algorithms**   In the context of deterministic planning, the *A$^*$ algorithm* is a heuristic search algorithm which is guaranteed to return optimal solutions when used with an admissible heuristic. A$^*$ extends breadth-first search and assigns each search node $n$, consisting of a state $s$ and a pointer to its parent node, an $f(n) = g(n) + h(s)$ value where $g(n)$ is the cost of the path from the initial state to $n$, and $h(s)$ is the heuristic estimate of the state. Search nodes are added and popped from a priority queue based on their $f$ values.

*Greedy Best First Search* (GBFS) and Weighted A$^*$ are variants of A$^*$ where $g(n) = 0$ and $f(n) = g(n) + w \cdot h(s)$ for some $w \in [1, \infty)$, respectively, for use in satisficing planning. GBFS has been extended in various ways for more efficient satisficing planning such as by including multiple-queues when using multiple heuristics, making use of preferred operators, and lazily deferring heuristic evaluation (Richter and Westphal 2010; Röger and Helmert 2010).

Anytime heuristic search (Hansen and Zhou 2007) refers to a class of heuristic search algorithms that continually return solutions of increasing quality, some of which converge to optimal solutions. Such algorithms are often able to use inadmissible heuristics which provide more informative $h^*$ estimates for better anytime behaviour.

AP researchers also use heuristic search algorithms for probabilistic planning. LAO$^*$ (Hansen and Zilberstein 2001) and LRTDP (Bonet and Geffner 2003) are optimal heuristic search algorithms with an explicit termination condition once an optimal solution is reached. We classify these algorithms as analytic algorithms as they require access to the transition function $\mathcal{T}$. Monte Carlo Tree Search algorithms such as UCT (Kocsis and Szepesvári 2006) are anytime optimal algorithms which can operate with only a generative transition model and use an inadmissible heuristic for anytime search, which we classify as sampling-based. However, sampling-based algorithms have a worst-case convergence rate significantly worse than analytic algorithms (Coquelin and Munos 2007; Walsh, Goschin, and Littman 2010). Regardless, algorithms have been proposed which combine the benefits of both worlds (Bonet and Geffner 2012; Keller and Helmert 2013; Wissow and Asai 2024). Heuristic search algorithms have also been extended to handle *constrained* MDPs (Trevizan et al. 2016; Trevizan, Thiébaux, and Haslum 2017; Schmalz and Trevizan 2024) by exploiting the convex, dual LP formulation of constrained MDPs combined with heuristics from the primal space.

**Problem Decomposition** So far, we have seen how structured representations offer a rich variety of methods for synthesising heuristic functions for use with heuristic search. Such representations also allow us to decompose hard, complex problems into easier and smaller subtasks. The idea of decomposition is also prevalent in HRL in its various forms (Dayan and Hinton 1992; Kaelbling 1993; Parr and Russell 1997; Sutton, Precup, and Singh 1999; Dietterich 2000). AP techniques described in this section take advantage of structured representations to compute decompositions without any search, learning or advice.

*Subgoals* can be viewed as one form of decomposition, indicating intermediate goals that must be achieved to synthesise a solution. They were initially introduced through the concept of landmarks (Porteous, Sebastia, and Hoffmann 2001; Richter, Helmert, and Westphal 2008) which may either be a fact that must be reached by any solution, or a set of actions of which at least one must exist in any solution. Subgoals have also been recently represented as sketches which can be used to solve specific domains in polynomial time (Bonet and Geffner 2021, 2024).

*Mutexes and invariants* represent sets of facts that are mutually exclusive in every state, and formulae that hold true in every state and goal condition, respectively. Mutexes are useful for transforming planning tasks into compact representations consisting of a small number of finite domain variables, rather than sets of facts as described previously (Helmert 2006, 2009). Invariants on the other hand can be used to prune irrelevant states and actions (Alcázar and Torralba 2015; Fiser and Komenda 2018). Such representations allow for more efficient problem analysis.

*Factored* planning (Brafman and Domshlak 2006) and *symbolic* search (Edelkamp 2003; Speck 2022) allow for performing search that does not incur exponential blowups in certain domains (Gnad 2021, Sec. 6). Factored planning refers to the idea of partitioning state variables for decomposing problems for which one can perform decoupled search (Gnad and Hoffmann 2018). Symbolic search refers to a class of algorithms performing reasoning or search over sets of states represented by Binary Decision Diagrams.

*Symmetries* in planning state spaces and transitions can also be computed and detected (Pochter, Zohar, and Rosenschein 2011; Shleyfman et al. 2015; Sievers et al. 2019) in order to prune large segments of the state space. The concept of symmetries can again be viewed as a form of problem decomposition as it involves collapsing equivalent subproblems to generate an easier task to solve.

**Incorporating prior knowledge** General AI agents aim to complete tasks which in some form or another are specified by humans, whether it be through reward functions in RL, or goal-conditioned models in AP. As pointed out and studied by (Booth et al. 2023) truly faithful reward functions are generally sparse and represent goal specifications, whereas in practice RL practitioners encode prior knowledge into the reward function to help RL agents learn faster. Contrarily, AP practitioners funnel knowledge into or on top of the existing planning model of the task via formal languages, for which tools exist for repairing incorrect models (Lin, Grastien, and Bercher 2023; Gragera et al. 2023).

Works for incorporating prior knowledge in planning generally involve employing formal languages. *Hierarchical Task Networks* (HTNs) (Erol, Hendler, and Nau 1994; Nau et al. 2003; Bercher, Alford, and Höller 2019) are used to model hierarchies and orderings of tasks. Variants of *temporal logics* have been used to express both state and action-centric temporal preferences for modelling domain control knowledge (Bacchus and Kabanza 2000; Bienvenu, Fritz, and McIlraith 2006; Baier et al. 2008). Reward machines (Icarte et al. 2018, 2022) employ finite state machines for modelling reward functions with the ability to express temporal properties and properties of regular languages.

*Logic programming* languages have been used in AP under the nomenclature of axioms (Thiébaux, Hoffmann, and Nebel 2005) for improving modelling and speeding up search (Ivankovic and Haslum 2015), defining valid tasks for a domain (Grundke, Röger, and Helmert 2024), specifying goals for RL (Agostinelli, Panta, and Khandelwal 2024), and directly encoding solution strategies for planning domains (Chen, Horčík, and Šír 2024).

### 3.5 Planning Resources

**Benchmarks** Planning researchers have been running the International Planning Competition[2] (IPC) for over two decades which exhibits a large and diverse set of benchmarks for both classical planning and more expressive extensions. Some of the planning domains used in previous IPCs also contain generators for automatically constructing new problems[3]. The PDDLGym[4] (Silver and Chitnis 2020) package automatically converts PDDL domains and problems into OpenAI Gym environments for use with RL.

**Planners** Fast Downward[5] (Helmert 2006) is a well-known AP system consisting of various powerful planner configurations for solving classical planning tasks. To find planners that handle more expressive AP formalisms covered in Sec. 3.3 or specific problem setups, one can refer to planners submitted to the various IPC tracks and relevant papers or ask in the planning slack channel[6].

**Further References** The planning.domains[7] initiative provides several AP resources and solvers as a free and open-source cloud service. Other useful open-source libraries include the Unified Planning library[8] which provides a Python interface to model and solve planning problems using existing classical, numeric, temporal, multi-agent, and hierarchical planners, and the Scickit-Decide library[9] which provides a framework for specifying and solving RL, AP, and scheduling problems. For textbook references, we refer to books by Ghallab, Nau, and Traverso (2004) and Geffner and Bonet (2013) for comprehensive introductions to AP. The AI book (Russell and Norvig 2020) covers planning as well as other related AI topics. The PDDL book (Haslum et al. 2019) provides a comprehensive introduction and reference to the AP language and its extensions.

---

[2] https://www.icaps-conference.org/competitions/
[3] https://github.com/AI-Planning/pddl-generators
[4] https://github.com/tomsilver/pddlgym
[5] https://github.com/aibasel/downward
[6] https://tinyurl.com/planning-community
[7] http://planning.domains/
[8] https://github.com/aiplan4eu/unified-planning
[9] https://airbus.github.io/scikit-decide/

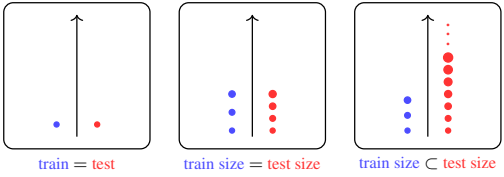

Figure 1: Generalisation setups for decision-making. AP approaches incorporating learning often handle the most general case (right) involving arbitrarily large test problems.

**AP in the Wild**  AP has also been used outside of benchmarks in the IPC. The 'poster-child of applications of AI planning' is the Mars Exploration Rovers (Bresina et al. 2005). More recently, AP technology has been incorporated into embodied AI systems (Liu, Palacios, and Muise 2023; Kumar et al. 2024). Common embodied AI benchmarks such as ALFWorld (Shridhar et al. 2021), a combination of the NLP TextWorld (Côté et al. 2019) and grounding ALFRED (Shridhar et al. 2020) benchmarks, use PDDL as a latent structure for representing the state of the environment. Conversely, common model-free RL benchmark domains such as MiniGrid (Chevalier-Boisvert et al. 2019) exhibit characteristics of classical planning problems: deterministic actions and goals in the form of sparse rewards. Such domains exhibit structured AP representations which can be explicitly encoded or learned (Sreedharan and Katz 2023).

## 4 Learning Structure and from Structure

RL setups involve learning from interactions with the environment for solving single tasks, but also learning from experiences that can be used for helping solve unseen tasks (Taylor and Stone 2009; Kirk et al. 2023). AP technology also exhibit learning capabilities, such as learning deadend detectors within a single problem (Steinmetz and Hoffmann 2017) analogous to clause-learning in SAT solvers, and computing pattern databases (Culberson and Schaeffer 1998) from abstractions of planning problems. In this section, we focus on learning and generalisation for AP across unseen problems. We begin by defining the generalisation problem for planning which is an out-of-distribution task. This is because evaluation is performed across unseen tasks with *arbitrary numbers of objects*.

This involves learning structured AP models from raw inputs which can generalise to unseen situations and scenarios which we cover in Sec. 4.1. By doing so, we can leverage AP technology for more efficient and effective decision-making as opposed to having no models of the world. Furthermore, we cover methodologies for learning directly from structured AP models in order to make planning faster. We treat this from two perspectives. In Sec. 4.2, we survey the statistical inference view of learning to plan from training data and the benefits of doing so from structured models. In Sec. 4.3, we cover program synthesis for constructing compact solutions of (possibly infinite) sets of structured AP problems with the added benefits of explainability and verifiability.

**Generalisation in Planning**
We tie together all aforementioned subfields of AP incorporating learning as the *Generalisation in Planning* setup, which we formulate and describe by its components:

1. A problem tuple $\langle \mathbf{D}, \mathbf{T}_{\text{train}}, \mathbf{T}_{\text{test}} \rangle$ where $\mathbf{D}$ is a domain, $\mathbf{T}_{\text{train}}$ is a finite set of training tasks drawn from $\mathbf{D}$, and $\mathbf{T}_{\text{test}}$ is a (possibly infinite) set of testing tasks from $\mathbf{D}$
2. An algorithm which outputs a set of plans corresponding to tasks in $\mathbf{T}_{\text{test}}$ and exhibits two modules:
   (a) a learner module, which takes in $\mathbf{D}$ and $\mathbf{T}_{\text{train}}$ and outputs a knowledge artifact, and
   (b) a planner module, which takes in arbitrary tasks from $\mathbf{T}_{\text{test}}$ and the artifact to output a solution for each task.

The Generalisation in Planning setup is similar to the Zero-Shot Generalisation in RL setup (Kirk et al. 2023) with the main difference lying in the specification of a domain, a set of similar tasks. The Zero-Shot RL setup formalises similarity of tasks with probability distributions in Contextual Markov Decision Processes which may not be known a priori. Conversely, the Generalisation in Planning setup takes advantage of formal languages in AP in order to explicitly define the domain and task.

### 4.1 Learning Planning Models

In this section, we see how MDPs at continuing, episodic, generative, or analytic access level can be *upgraded* to structured access by learning a transition model of the environment corresponding to $\mathcal{T}$. Model-based RL (MBRL) approaches sometimes rely on observations or active exploration of the environment to learn a model of the environment. The classic "Dyna" framework (Sutton 1991) in its introduction mentioned a step for "Learning of domain knowledge in the form of an action model", before using the learned model to "plan". A similar approach can be used to learn the domain model $\mathbf{D}$ in various forms. This model $\mathbf{D}$ compactly represents the set of transitions $\mathcal{T}$. Advantages of MBRL include better explainability of the decisions, sample efficiency and thereby better learning speed, and safety assessment of the agent (Moerland et al. 2023). Additionally, learning a domain model is better than learning a task-specific model (as done in many RL settings) because the learned domain model can be reused for any task following the same dynamics, irrespective of the number of objects, configuration of objects in the environment, etc.

**Learning from passive state-action traces**  Action Model Learning for Planning has a rich history of learning action models from a set of input state-action transitions (Arora et al. 2018; Aineto, Jiménez, and Onaindia 2022). The classical approaches learned PDDL-like models domain for deterministic transition systems. LOCM (Cresswell, Mc-Cluskey, and West 2009), LOCM2 (Cresswell and Gregory

2011), etc. present a class of algorithms that use finite-state machines to create PDDL models from observed plan traces. ARMS (Yang, Wu, and Jiang 2007), AMAN (Zhuo and Kambhampati 2013), etc. leverage MAX-SAT to learn action models with partial or noisy traces. FAMA (Aineto, Celorrio, and Onaindia 2019) reduces model recognition to a planning problem and can work with partial action sequences and/or state traces as long as correct initial and goal states are provided. Bonet and Geffner (2020a) and Rodriguez et al. (2021) present approaches for learning relational models using a SAT-based method when the action schema, predicates, etc. are not available. These approaches take as input a predesigned correct and complete directed graph encoding the structure of the entire state space. A central theme of these approaches is to analyse the states before (and after) an action is applied to learn the common features as preconditions (and effects). Callanan et al. (2022) provides APIs for many of these approaches.

**Learning from passive image traces** The works mentioned above learn a model of the agent using transitions that are represented using successive states in $\mathcal{S}$. Many times such information is not directly available, and there are approaches that learn the environment model using images. Asai et al. (2022); Verma, Marpally, and Srivastava (2021); Xi, Gould, and Thiébaux (2024), etc. learn lifted PDDL action models from image traces. These approaches have been shown to work well on domains in grid formats or from the PDDLGym library (Silver and Chitnis 2020). Campari et al. (2022) learn similar models for embodied AI domains using images of the environment to get state information.

**Learning from directed exploration** Xu and Laird (2010) and Lamanna et al. (2021) use online learning to learn a PDDL action model incrementally, instead of learning from observations. The idea is to incorporate new observations to improve the action model. This setting is closer to the typical MBRL setting. IRALe (Rodrigues et al. 2011) is an active learning based method that learns lifted transition modules by exploring actions in states where its partially learned preconditions almost hold. Verma, Marpally, and Srivastava (2021) use active querying to learn a PDDL model for an agent. Verma, Marpally, and Srivastava (2022) learn high level capabilities (or skills) of the agent and learn a PDDL model in terms of those capabilities.

Many recent approaches learn models for environment with stochastic transition systems. Incremental Learning Model (Ng and Petrick 2019) employs reinforcement learning to construct PPDDL models without relying on plan traces. This approach strategically prioritizes the execution of actions about which the model has limited information, gradually building a comprehensive PPDDL model of the environment through iterative learning. Chitnis et al. (2021) present an approach for learning probabilistic relational models where they use goal sampling as a heuristic for generating relevant data. The central idea for the agent is to generate a goal far away from the initial state, and the agent tries to reach that goal using the model known at that point. The data generated from executing the actions while trying to reach that goal is used to learn the environment model.

Verma, Karia, and Srivastava (2023) learn a PPDDL model using autonomous active querying. These queries are in the form of policies that the agent must execute in the environment, and learn a model based on the resulting observations.

### 4.2 Learning for Planning

As mentioned previously, MBRL exhibits various benefits from being able to access closed-form representations of MDP components, including data efficiency, safety, and explanability. In this section, we highlight further advantages for learning that can be gained from planning representations discussed in Sec. 3. Notably, such representations exhibit rich structural information to aid learning by improving efficiency and performance, and do not require additional computation to learn from scratch.

Before we begin, we also make special mention to related fields of Relational RL (Dzeroski, Raedt, and Driessens 2001; Hazra and Raedt 2023; Marra et al. 2024) and Inductive Logic Programming (Muggleton and Raedt 1994; Lavrac and Dzeroski 1994; Cropper et al. 2022) which similarly learn from structured data using first-order logic.

**Low cost of reasoning** Advantages of planning representations for learning to solve decision making tasks are manifold arising from the rich structural information of the task at hand they implicitly encode. One advantage of this is the low number of data samples required to learn domain knowledge for unseen tasks. In comparison to model-free approaches which have to learn latent features from both raw inputs and exploration, planning representations operate on relational information from which generating features is efficient and well-aligned with the task at hand.

Early works in L4P exploited the structure of planning tasks to quickly and automatically derive features (Martín and Geffner 2004; Yoon, Fern, and Givan 2008; Celorrio, Aguas, and Jonsson 2019) or logical rules (Khardon 1999; Gretton and Thiébaux 2004) that were correlated with and relevant to the task at hand. Despite the much more primitive compute available at the time of such works, these methods were able to solve simple planning tasks such as Blocksworld which are still unsolved with deep learning and large architectures at the time this paper was written (Valmeekam et al. 2023b,a; Valmeekam, Stechly, and Kambhampati 2024). Furthermore, recent works incorporating learning in planning continue to make use of small datasets, with an emphasis of demonstrating high data efficiency. For example, experiments in (Chen and Thiébaux 2024) use a few dozen small training tasks and a median time of one second to generate labels for use in supervised learning across various domains.

**Exploiting graph learning research** More recent research progress in L4P has been exploiting graph learning architectures such as message passing graph neural networks (GNNs) (Gilmer et al. 2017), in line wth the current surge of interest in the latter.[10] Graph learning approaches can take on arbitrarily sized inputs, display extrapolation capabilities (Xu et al. 2021) and theoretical relations to counting logics (Morris et al. 2019; Xu et al. 2019; Barceló et al. 2020;

---
[10]Based on ICLR'24 publications: https://tinyurl.com/3xjknfhh

Grohe 2021) that standard neural networks operating on raw inputs do not exhibit.

L4P graph learning works generally learn reusable heuristic and value functions (Shen, Trevizan, and Thiébaux 2020; Ståhlberg, Bonet, and Geffner 2022; Chen, Thiébaux, and Trevizan 2024), policies (Toyer et al. 2018, 2020; Dong et al. 2019; Garg, Bajpai, and Mausam 2020; Sharma et al. 2022; Wang and Thiébaux 2024), or preprocessing techniques (Ma et al. 2020; Silver et al. 2021) that can be used to solve unseen tasks. Furthermore and in line with the previous point of low cost of reasoning, it has been shown that classical machine learning making use of graph kernels can be used to train models as expressive as GNNs and solve AP tasks several orders of magnitude faster than deep learning (Chen, Trevizan, and Thiébaux 2024; Chen and Thiébaux 2024).

**Automatic reward functions and labels** Standardised representations of planning tasks allow for various methods of automatically generating faithful reward functions or supervised training labels for arbitrary domains and tasks. Related to the previous point, this means that there is no need for planning practitioners to design reward functions for each domain individually and instead it is possible to take off-the-shelf algorithms for generating rewards and labels for learning. For example, the ASNets (Toyer et al. 2018, 2020) framework and its numeric planning extension (Wang and Thiébaux 2024) performs an RL-style approach of exploring with a partially learned policy and exploiting with a *teacher planner* as a proxy for a reward function. It is also possible to exploit various existing domain-independent heuristics (Bonet and Geffner 2001) that are commonly used in planning as dense reward functions for RL approaches (Gehring et al. 2022), and conversely use RL directly for learning heuristic functions (Micheli and Valentini 2021; Ståhlberg, Bonet, and Geffner 2023).

Furthermore, it is possible to automatically generate supervised training labels from training tasks with domain-independent planners directly. This is can be done for learning heuristic functions by extracting optimal value functions from optimal plan traces. Works have also exploited the fact that in certain search algorithms, heuristics are better viewed as *ranking functions* (Garrett, Kaelbling, and Lozano-Pérez 2016), for which it is also possible to learn from optimal plans with various different optimisation formulations using RankSVM (Garrett, Kaelbling, and Lozano-Pérez 2016), differentiable loss functions (Chrestien et al. 2023), classification (Hao et al. 2024), and Linear Programs (Chen and Thiébaux 2024). Orthogonally, Núñez-Molina et al. (2024) proposed improved regression functions for learning heuristics using bounds derived from truncated Gaussians.

### 4.3 Generalised Planning
This area of research focuses on computing and representing generalisable knowledge for planning. In the context of the *Generalisation in Planning* framework articulated at the start of Sec. 4, this class of approaches addresses the problem of computing or learning the "knowledge artifact" or Generalised Planning Knowledge (GPK) in the form of algorithmic generalised plans (Levesque 2005; Srivastava, Immerman, and Zilberstein 2008; Bonet, Palacios, and Geffner 2009; Srivastava, Immerman, and Zilberstein 2011; Hu and Giacomo 2011; Segovia-Aguas, Jiménez, and Jonsson 2016; Silver et al. 2024), generalised policies and Q-functions for stochastic settings (Fern, Yoon, and Givan 2006; Ng and Petrick 2022; Karia, Kaur Nayyar, and Srivastava 2022) and other forms of knowledge for reducing the computational complexity of future planning tasks. Most work in this area focuses on learning via program synthesis for transfer to tasks with unbounded numbers of objects; almost all work addresses scenarios where test tasks feature more objects (and corresponding longer horizons and larger, different state spaces) than training tasks.

Approaches for Generalised Planning (GP) can be evaluated on various dimensions (Srivastava, Immerman, and Zilberstein 2011), including the *domain coverage*, i.e., the fraction of tasks of interest where it would be applicable; and *cost of instantiation* or computational complexity of using GPK to compute solutions for new tasks in a zero-shot fashion. Typically, these two metrics are inversely correlated. Learning generalised plans is desirable because they yield high domain coverage at low costs of instantiation. However, finding such generalised solutions is challenging: it is well known that relative simple representations for generalised plans and are equivalent to counter-based models of computing, and thus equivalent to Turing equivalent, making it difficult to determine, in general, whether a given generalised plan will terminate and achieve a goal condition.

Several algorithms push the boundary of decidability on this problem for structured classes of generalised plans (Srivastava 2010; Hu and Giacomo 2011; Srivastava, Immerman, and Zilberstein 2012). The sieve family of algorithms (Srivastava et al. 2011; Bonet and Geffner 2020b; Srivastava 2023) presents structure-independent, sound approaches for evaluating counter-based generalised plans. In conjunction with logic-based abstractions that map problem states into vectors of counters, these algorithms have been used to compute generalised plans as well as sketches (Bonet and Geffner 2021), as another form of GPK.

As in approaches for learning for planning, multiple classes of models have been considered: generative models (simulators) as well as analytic and structured models. While much of the work discussed above utilizes structured domain models, recent approaches also address the problem of learning GPK without analytical domain models. In particular, recent research indicates that deep learning in conjunction with logic-based abstractions originally developed for learning generalised plans yields sample efficient approaches for learning generalised heuristics (e.g., (Karia and Srivastava 2021; Ståhlberg, Bonet, and Geffner 2022; Chen, Thiébaux, and Trevizan 2024)) and generalised Q-functions in RL problems (Karia and Srivastava 2022) without access to analytic/structured domain models.

## 5 Conclusion
This survey paper provides a primer into the field of AI Planning (AP) by introducing core concepts and ideas revolving around the theme of *structure*. Furthermore, we covered areas concerned with learning structure in order to perform AP, as well as learning from structure for improving AP.

## Acknowledgements
The authors would like to thank Tom Silver, Sheila McIlraith, and Scott Sanner for many helpful discussions and feedback on this survey.